\documentclass[twoside,leqno,twocolumn]{article}

\usepackage[letterpaper]{geometry}

\usepackage{ltexpprt}
\usepackage{hyperref}
\usepackage{graphicx}
\usepackage{booktabs}
\usepackage{amsmath}
\begin{document}

\newcommand\relatedversion{}

\title{\Large 
CISum: Learning Cross-modality Interaction to Enhance Multimodal Semantic Coverage for Multimodal Summarization}
\author{Litian Zhang\textsuperscript{\dag}
\and Xiaoming Zhang\textsuperscript{\dag}\thanks{Corresponding author.}
\and Ziming Guo\textsuperscript{\dag}
\and Zhipeng Liu\thanks{School of Cyber Science and Technology, Beihang University Email:\{litianzhang, yolixs, guoziming, lzpeng\}@buaa.edu.cn}
}

\date{}

\maketitle


\fancyfoot[R]{\scriptsize{Copyright \textcopyright\ 2023 by SIAM\\
Unauthorized reproduction of this article is prohibited}}





\begin{abstract} \small\baselineskip=9pt Multimodal summarization (MS) aims to generate a summary from multimodal input. 
Previous works mainly focus on textual semantic coverage metrics such as ROUGE, which considers the visual content as supplemental data. Therefore, the summary is ineffective to cover the semantics of different modalities. 
This paper proposes a multi-task cross-modality learning framework (CISum) to improve multimodal semantic coverage by learning the cross-modality interaction in the multimodal article.  
To obtain the visual semantics, we translate images into visual descriptions based on the correlation with text content. 
Then, the visual description and text content are fused to generate the textual summary to capture the semantics of the multimodal content, and the most relevant image is selected as the visual summary. 
Furthermore, we design an automatic multimodal semantics coverage metric to evaluate the performance. 
Experimental results show that CISum outperforms baselines in multimodal semantics coverage metrics while maintaining the excellent performance of ROUGE and BLEU.
\end{abstract}
\section{Introduction}

Multimodal summarization (MS) condenses the multimodal news article into a short summary, which helps users grasp crucial information quickly.
One of the core problems of summarization is to ensure the summary covers the facts of the source article as more as possible. 
As for the text-only summarization, many works \cite{see2017get,Cohan2018ADA,Chowdhury2020NeuralAS} aim to improve the word overlap metric, such as ROUGE. 
Similar to these works, most of the MS models also focus on improving the overlap toward the source text only \cite{li2018multi,palaskar2019multimodal,li2020multimodal}, while the visual content is considered as the supplemental data handled independently. However, the visual content may convey some other information besides the textual content.
Therefore, the summary isn't effective to cover the facts of both the textual and visual content in the source article, i.e., low coverage of multimodal semantics.

\begin{figure}[!t]
\setlength{\abovecaptionskip}{0.1cm} 
\setlength{\belowcaptionskip}{-0.5cm} 
\begin{center}
\includegraphics [scale=0.56]{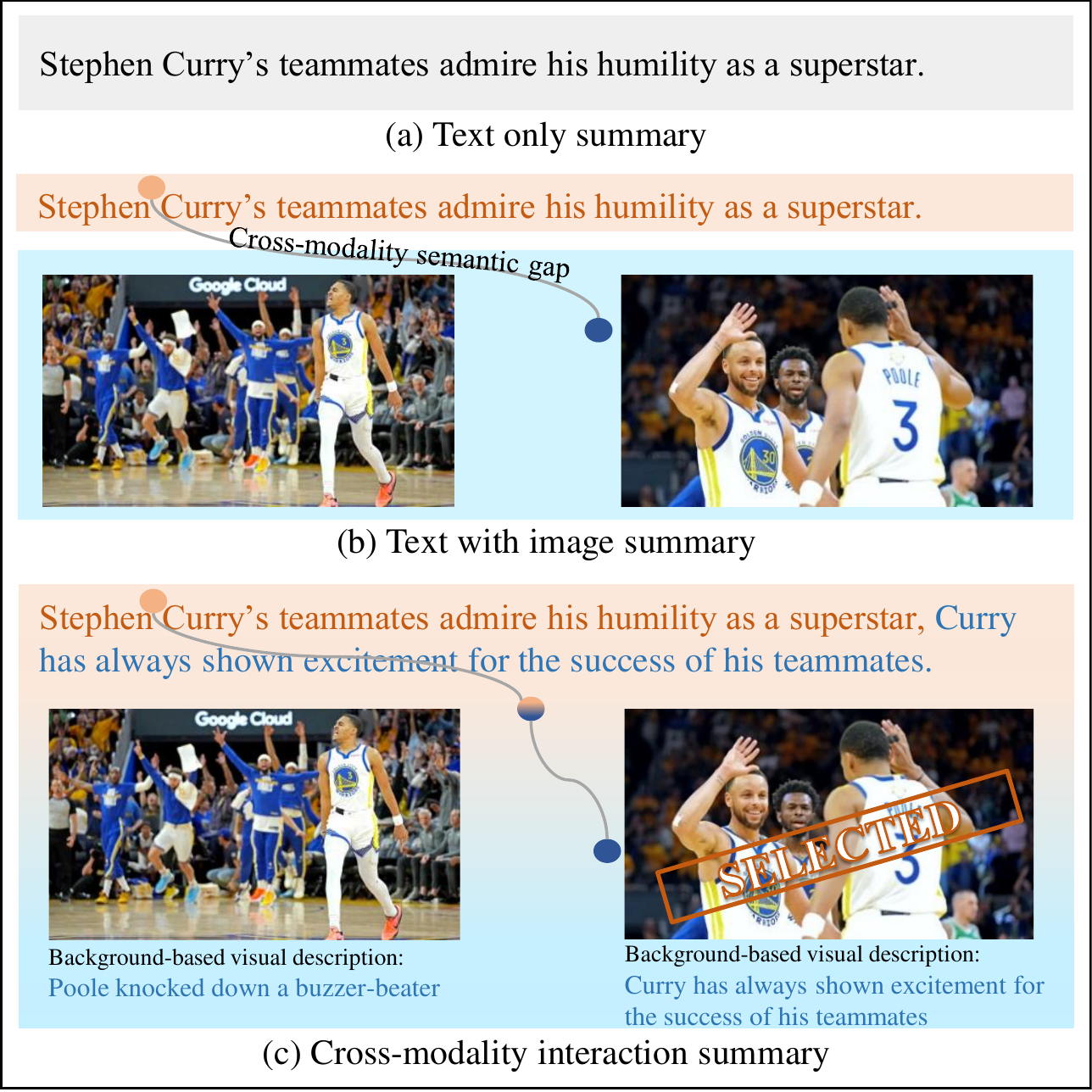}
\end{center}
\caption{Comparison of the three summary output results in multimodal summarization task.}
\label{fig:3example}
\end{figure}

As shown in Figure \ref{fig:3example} (a), the text-only summary only reflects the semantics of the text content, which cannot cover the visual semantics when it is different from the text.  
One direct way to enhance visual semantics coverage is to extract the most text-relevant image and generate a multimodal summary. 
For example, \cite{li2018multi} points out that visual information can improve user satisfaction and propose an attention-based model to generate the summary containing an image. \cite{li2020vmsmo} also proposes a multimodal summarizer to generate the text summary and choose the most relevant image. However, the two types of content are handled independently, where the cross-modality interaction cannot be effectively captured. As Figure \ref{fig:3example} (b) shows, a contextually irrelevant image may cause ambition and confuse readers. Directly combining the textual and visual information may also cause cross-modality inconsistency, which accordingly affects the multimodal semantics coverage.

Compared with the textual feature, the visual feature is a much lower level representation of semantics \cite{Liu2014PersonalizedGT}. We argue that the visual content should be transformed into a high-level semantic space corresponding with the textual content, which can decrease the semantic gap between the two modalities. Meanwhile, the cross-modality interaction should be effectively learned to fuse different modalities of content as a whole to generate the summary and improve multimodal semantics coverage. 
To obtain the high-level semantic representation, we learn the article's background-based visual description. 
Unlike image caption, the background-based visual description is obtained by translating image content into text description based on the source article background.
As shown in Figure \ref{fig:3example} (c), the visual description reflects the semantics of the image and is more effective to learn the cross-modality interaction with the textual content, and then improves multimodal semantics coverage.

However, it is a great challenge to simultaneously generate the background-based visual description and multimodal summary.
First, both the visual description and textual summary should capture the crucial information of the multimodal article but focus on different modalities.
It is nontrivial to learn the multimodal representation effectively to process the two-generation tasks simultaneously. 
Second, the visual description generation is based on the relevant image and parts of textual content. However, there is no obvious knowledge about this information, and much noisy information may be included. 
Third, measuring multimodal semantics and cross-modality semantic similarity are rarely studied before.

To tackle these challenges, we propose a multi-task cross-modality learning framework to generate cross-modality interaction summaries (CISum). 
In particular, three tasks for MS are defined, i.e., visual-aware summary generation, relevant image selection, and visual description generation. CISum optimizes the three tasks simultaneously to improve multimodal semantics coverage of the generated summary. A noisy filter cross-modality attention module is proposed to reduce the noisy influence, which helps the model to learn the cross-modality interaction.
Furthermore, multimodal semantics coverage metrics are designed to evaluate the cross-modality semantics similarity, which is more reasonable for multimodal summarization than textual metrics. 
Experimental result shows that CISum can improve the multimodal semantics coverage and is also excellent on word overlap. The contribution can be concluded as follows:

\begin{itemize}
\item To the best of our knowledge, this is the first work that proposes to explore the problem of multimodal semantics coverage in MS.

\item We propose a novel model to improve multimodal semantics coverage, which learns the cross-modality interaction based on a multi-task cross-modality learning framework. 

\item We design multimodal semantics coverage metrics to evaluate MS, and the experiments demonstrate the superiority of CISum on these metrics and overlap words metrics.
\end{itemize}

\section{Problem Formulation}

\begin{figure*}[!t]
\setlength{\abovecaptionskip}{0.05cm} 
\setlength{\belowcaptionskip}{-0.3cm} 
\begin{center}\includegraphics [scale=0.52]{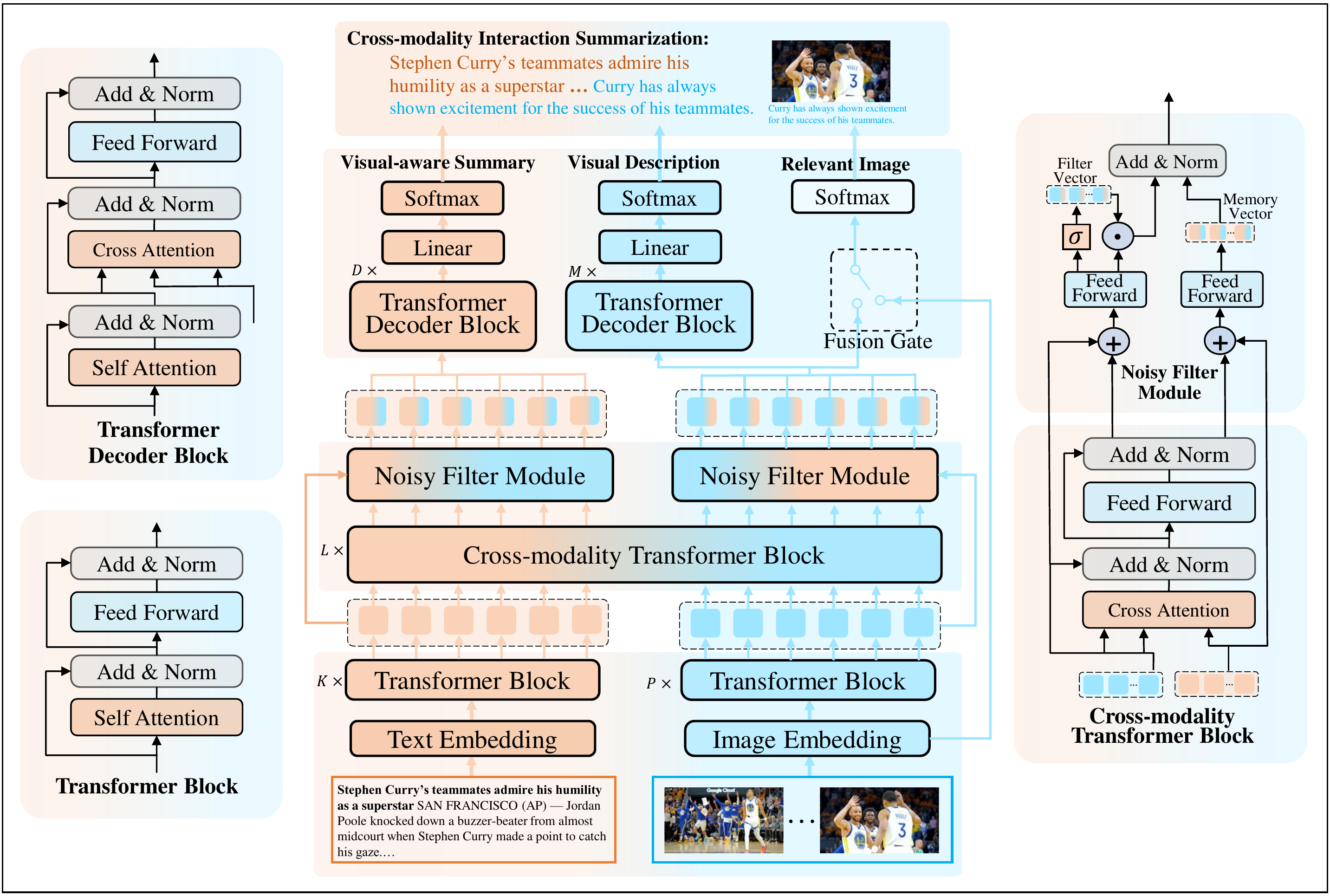}
\end{center}
\caption{The overview of CISum.}
\label{fig:model}
\end{figure*}

Given an input news article with the textual content containing $N$ words $X_t = \{x_1, x_2, \dots , x_N\}$ and the visual content containing $M$ images $X_v = \{i_1, i_2, \dots, i_M\}$, CISum obtains three parts of output by three tasks, i.e., visual-aware summary $Y_t = \{y_1, y_2, \dots , y_P\}$ with $P$ words, most relevant image $Y_v = \{i_m\}$ and visual description $Y_d = \{Y_d^1,Y_d^2,\dots,Y_d^M\}$ for $M$ input images. Then, CISum can be formulated as an optimization problem as follows:
$$
\setlength{\abovedisplayskip}{3pt}
\setlength{\belowdisplayskip}{3pt}
\arg\max_{\theta} CISum(Y_t,Y_v,Y_d|X_t,X_v;\theta)
$$
To obtain the multimodal summary $Y_{s} = \{y_1,y_2, \dots,Y_d^m,\dots,y_P \}$, a visual description locator is proposed to determine the location in textual summary $Y_t$ to insert the relevant image description $Y_d^m$. Then, the Cross-modality Interaction Summary is defined as $(Y_{s},Y_v)$, which indicates a cross-modality semantic  textual summary associate with the most relevant image.

\section{Our Model}
The current multimodal summarization methods mainly focus on supplementing the text features with visual information to improve word overlap metrics, which is not effective to improve multimodal semantics coverage. 
Therefore, we propose a novel multi-task cross-modality learning framework to learn the cross-modality interaction to generate the summary. As shown in Figure 2, CISum comprises three modules, i.e.,  Feature Embedding used to encode each modality into feature vectors, Cross-modality Interaction Encoder used to learn cross-modality correlation and filter the unrelated noisy information, and Multi-task Triple Decoder used to generate the visual-aware summary, visual description, and relevant image. 
Meanwhile, an unsupervised trained visual description locator is proposed to insert the visual description into the textual summary.

\subsection{Feature Embedding}
As shown in Figure \ref{fig:model}, the textual and visual inputs are embedded into vector representations firstly. We apply the pretrained BERT \cite{2018BERT} model to embed the input text $X_t = \{x_1, x_2, \dots , x_N\}$ into the embedding of words $ H_t = \{h_1, h_2, \dots , h_N\}$. As for the visual input $X_v = \{i_1, i_2, \dots, i_M\}$,  we apply a Faster R-CNN \cite{ren2016faster}) initialized with ResNet-101 \cite{he2016deep} to generated the visual representations $ G = \{v_1, v_2, \dots , v_M\}$.

Then a series of transformer blocks (TRM) are applied to encode the initial feature. The $j$-th layer TRM is defined as follows:
\begin{equation}
\setlength{\abovedisplayskip}{6pt}
\setlength{\belowdisplayskip}{6pt}
\begin{split}
    \widetilde{H}^{j-1} &= \text{MAtt}(Q{H}^{j-1},K{H}^{j-1},V{H}^{j-1}) \\
    \widetilde{H}^{j} &= \text{LN}({H}^{j-1}+\widetilde{H}^{j-1}) \\
    H^{j} &= \text{LN}(\widetilde{H}^{j}+\text{FFN}(\widetilde{H}^{j})) \\
\end{split}
\end{equation}
where $H$ is the input vector, LN is the layer normalization, MAtt is the multihead attention, FFN is the feed forward normalization, the parameters $Q\in {R}^{d_{e}\times d_h}$, $K\in {R}^{d_{e}\times d_h}$ and $V\in {R}^{d_{model}\times d_h}$ are the query, key and valus matrices, $d_h$ denotes the dimension of hidden representation, and $d_{e}$ denotes the dimension of attention embedding. The text and visual features are then encoded as follows:
\begin{equation}
\setlength{\abovedisplayskip}{4pt}
\setlength{\belowdisplayskip}{6pt}
\begin{split}
    \hat{H}_t &= \text{TRM}(H_t^{K-1},H_t^{K-1})\\
    \hat{G} &= \text{TRM}(G^{P-1},G^{P-1})\\
\end{split}
\end{equation}
where $\hat{H}_t$ is the textual output of the $K^{th}$ layer of TRM, and $\hat{G}$ is the visual output  of the $P^{th}$ layers of TRM.

\subsection{Cross-modality Interaction Encoder}
To improve the multimodal semantics coverage, different modalities should be fused seamlessly to generate the summary by exploiting the interaction between them. Usually, there exists redundancy information in different modalities, and the unrelated information may also affect the fusion of the contents. 
Especially for the visual content, only a small part of the text semantics are correlated with the visual semantics.
Therefore, a noisy filter module is proposed to capture the crucial cross-modality correlation information during feature fusion.

We first apply the cross-modality transformer block (Co-TRM) after the feature embedding to learn the cross-modality fusion feature. The $j$-th layer of Co-TRM is defined as follows:
\begin{equation}
\setlength{\abovedisplayskip}{6pt}
\setlength{\belowdisplayskip}{6pt}
\begin{split}
    \widetilde{G}^{j-1} &= \text{MAtt}(Q{H}^{j-1},K{G}^{j-1},V{G}^{j-1}) \\
    \widetilde{G}^{j} &= \text{LN}({H}^{j-1}+\widetilde{H}^{j-1}) \\
    G^{j} &= \text{LN}(\widetilde{H}^{j}+\text{FFN}(\widetilde{H}^{j})) \\
\end{split}
\end{equation}
where $H$ is the textual vector, and $G$ is the visual vector. Meanwhile, a noisy filter mechanism is used to learn the crucial cross-modality information, which aims to reduce the noisy influence. As shown on the right of Figure \ref{fig:model}, the noisy filter process on the textual modality is as follows:
\begin{equation}
\setlength{\abovedisplayskip}{6pt}
\setlength{\belowdisplayskip}{6pt}
\begin{split}
H_{cross}^{L-1} &= \text{Co-TRM}(\hat{H}_t^{L-1},\hat{G}^{L-1})\\
\hat{G}^{L-1}_{memory} &= \text{FFN}(concat(H_{cross}^{L-1},\hat{G}^{L-1}))\\
\hat{G}^{L-1}_{filter} &= sigmod(\hat{G}^{L-1}_{memory})\\
\end{split}
\end{equation}
where $H_{cross}^{L-1}$ is the textual fusion vector, $\hat{G}^{L-1}_{memory}$ is the visual memory vector, and $\hat{G}^{L-1}_{filter}$ is the visual filter vector gate to filter noisy from visual modality. Then the filtered visual information and textual memory are fused again:
\begin{equation}
\setlength{\abovedisplayskip}{6pt}
\setlength{\belowdisplayskip}{3pt}
\begin{split}
\hat{H}^{L-1}_{memory} &= \text{FFN}(concat(H_{cross}^{L-1},\hat{H}_t^{L-1}))\\
H_{enc} &= \text{LN}(\hat{G}^{L-1}_{filter}\odot \hat{G}^{L-1}_{memory}+\hat{H}^{L-1}_{memory})
\end{split}
\end{equation}
where $\hat{H}^{L-1}_{memory}$ is the textual memory vector, and $H_{enc}^{L-1}$ is the textual vector after $L$ stack layers of filter noisy filter cross-modality fusion. Besides the visual encoder, $\hat{G}^{L-1}$ is also processed by a noisy filter mechanism, and $G_{enc}$ is the visual vector after $L$ stack layers.

\subsection{Multi-task Triple Decoder}
Similar to the text-only summarization, some existing MS approaches only contain a textual decoder without the visual information output. Other approaches include an image selector to generate multimodal output, which is not effective to learn the cross-modality interaction. As shown in Figure \ref{fig:model}, we propose a Multi-task Triple Decoder to divide the multimodal output into three parts based on cross-modality interaction. This module receives cross-modality fusion vectors as the input for the three decoder parts, which generates text summary, selects relevant images, and generates visual descriptions based on the cross-modality semantic correlation, respectively.

\subsubsection{Visual-aware Summary Generation}
This module utilizes transformer decoder block (De-TRM) to decode vectors into output words. The $j$-th layer of De-TRM is defined as follows:
\begin{small}
\begin{equation}
\setlength{\abovedisplayskip}{6pt}
\setlength{\belowdisplayskip}{6pt}
\begin{split}
        \widetilde{H}_s^{j} &= \text{LN}({D}_w^{j-1}+\text{MAtt}(Q_0{H_s}^{j-1},K_0{H_s}^{j-1},V_0{H_s}^{j-1}))\\
        \widetilde{H}_{enc}^{j} &= \text{LN}(\widetilde{H}_s^{j}+\text{MAtt}(Q_1\widetilde{H}_s^{j},K_1H_{enc},V_1H_{enc}))\\
        H_{enc}^{j} &= \text{LN}(\widetilde{H}_{enc}^{j}
    +\text{FFN}(\widetilde{H}_{enc}^{j}))
\end{split}
\end{equation}
\end{small}
where $H_s^{j-1}$ is the previous $s$ token embeddings to calculate the vocabulary distribution of the next token, the visual-aware summary is generated as:
\begin{equation}
\setlength{\abovedisplayskip}{6pt}
\setlength{\belowdisplayskip}{6pt}
\begin{split}
{H}_s &= \text{De-TRM}(H_s^{D-1},H_{enc}^{D-1})
\end{split}
\end{equation}
where $H_s$ is token embedding vector after $D$ layers of De-TRM, the output of all  tokens is ${H}_s = \{{h}_1, {h}_2, \dots, {h}_S\}$. Then,  the vocabulary distribution is obtained based on ${h}_s$ as follows:
\begin{equation}
\setlength{\abovedisplayskip}{6pt}
\setlength{\belowdisplayskip}{6pt}
\begin{split}
p_{s} = {{softmax}}(\text{FFN}({W}{h}_s))
\end{split}
\end{equation}
where ${W}$ is a trainable parameter matrix. $p_{s}$ is used to calculate the cross-entropy loss:
\begin{equation}
\setlength{\abovedisplayskip}{6pt}
\setlength{\belowdisplayskip}{6pt}
\begin{split}
\mathcal{L}_{text}=-\sum_{s=1}^{S} \boldsymbol{y}_{s}^{T} \log \left(\boldsymbol{p}_{s}\right)
\end{split}
\end{equation}

\subsubsection{Visual Description Generation}
Unlike image caption, the background-based visual description aims to translate image content into text description based on the background information of the news article. Usually, the textual content in the source article contains a great part of information unrelated to the images. The noisy filter module is applied to fuse visual semantic features with the correlated text features. Given the filtered visual vector $G_{enc}$ and textual vector $H_{enc}$, the visual description is generated as follows:
\begin{equation}
\setlength{\abovedisplayskip}{6pt}
\setlength{\belowdisplayskip}{6pt}
\begin{split}
G_{f} &= \text{Co-TRM}(G_{enc},H_{enc})\\
G_{d} &= \text{De-TRM}(G_{d}^{M-1},G_{f}^{M-1})\\
p_{d} &= {{softmax}}(\text{FFN}({W'}{G}_d))
\end{split}
\end{equation}
where $G_{f}$ is the fusion of visual and textual vectors, $G_{d}$ is the token embedding after $M$ layers of De-TRM, the output of all tokens is ${G}_d = \{{g}_1, {g}_2, \dots, {g}_D\}$, $W'$ is a trainable parameter matrix for the vocabulary distribution, and $p'_{d}$ is vocabulary distribution. The visual description cross-entropy loss is defined as follows:
\begin{equation}
\setlength{\abovedisplayskip}{3pt}
\setlength{\belowdisplayskip}{3pt}
\begin{split}
\mathcal{L}_{visual}=-\sum_{d=1}^{D} \boldsymbol{y}_{d}^{T} \log \left(\boldsymbol{p}_{d}\right)
\end{split}
\end{equation}
\subsubsection{Relevant Image Selection}
The most relevant image should match the generated summary. Therefore, this image is selected based on the cross-modality fusion visual vector  $G_{enc}$ and image embedding  $G$. First, the image embedding vector $G$ is used to calculate the fusion weight score. Then, the fusion gate fuses the visual vector $G_{enc}$ with $G$ and obtains the image ranking scores. The process can be formulated as follows:

\begin{equation}
\setlength{\abovedisplayskip}{2pt}
\setlength{\belowdisplayskip}{3pt}
\begin{split}
\lambda &=\sigma(\text{FFN}(G))\\
\hat{I} &= \lambda G+(1-\lambda)G_{enc}\\
I_{score} &=  softmax(\text{FFN}(\hat{I}))\\
\mathcal{L}_{image}&=-\boldsymbol{y}_{I}^{T} \log \left(\boldsymbol{I}_{score}\right)
\end{split}
\end{equation}
where $ \lambda $ and $\lambda^{*}$ are the balance weights, $\sigma$ is the activation function, $I_{score}$ is the images ranking score, and $\mathcal{L}_{image}$ denotes  the cross-entropy loss of image selection.

We also apply a dynamic weight averaging mechanism \cite{Liu2019EndToEndML} for multi-task loss to ensure that each task is learned at a similar rate. The sub-task weight is calculated as:
\begin{equation}
\setlength{\abovedisplayskip}{3pt}
\setlength{\belowdisplayskip}{3pt}
\begin{split}
\quad w_{k}(t-1)&=\frac{\mathcal{L}_{k}(t-1)}{\mathcal{L}_{k}(t-2)}\\
\lambda_{k}(t)&=\frac{K \exp \left(w_{k}(t-1) / T\right)}{\sum_{i} \exp \left(w_{i}(t-1) / T\right)}\\
\end{split}
\end{equation}
where $\mathcal{L}_{k}(t-1)$ is the $k^{th}$ task loss value at step $t-1$, $\quad w_{k}(t-1)$ is the trained loss and $\lambda_{k}(t)$ is the weight. $K$ is the number of tasks, and here $K=3$. Then, the global loss at step  $t$ is:
\begin{equation}
\setlength{\abovedisplayskip}{3pt}
\setlength{\belowdisplayskip}{3pt}
\begin{split}
\mathcal{L}_{global }(t) =\lambda_{text}\mathcal{L}_{text}(t)+\lambda_{visual}\mathcal{L}_{visual}(t)\\+\lambda_{image}\mathcal{L}_{image}(t)
\end{split}
\end{equation}

\subsection{Visual Description Locator}

\begin{figure}[!t]
\setlength{\abovecaptionskip}{0.05cm} 
\setlength{\belowcaptionskip}{-0.6cm} 
    \begin{center}
    \includegraphics[scale=0.60]{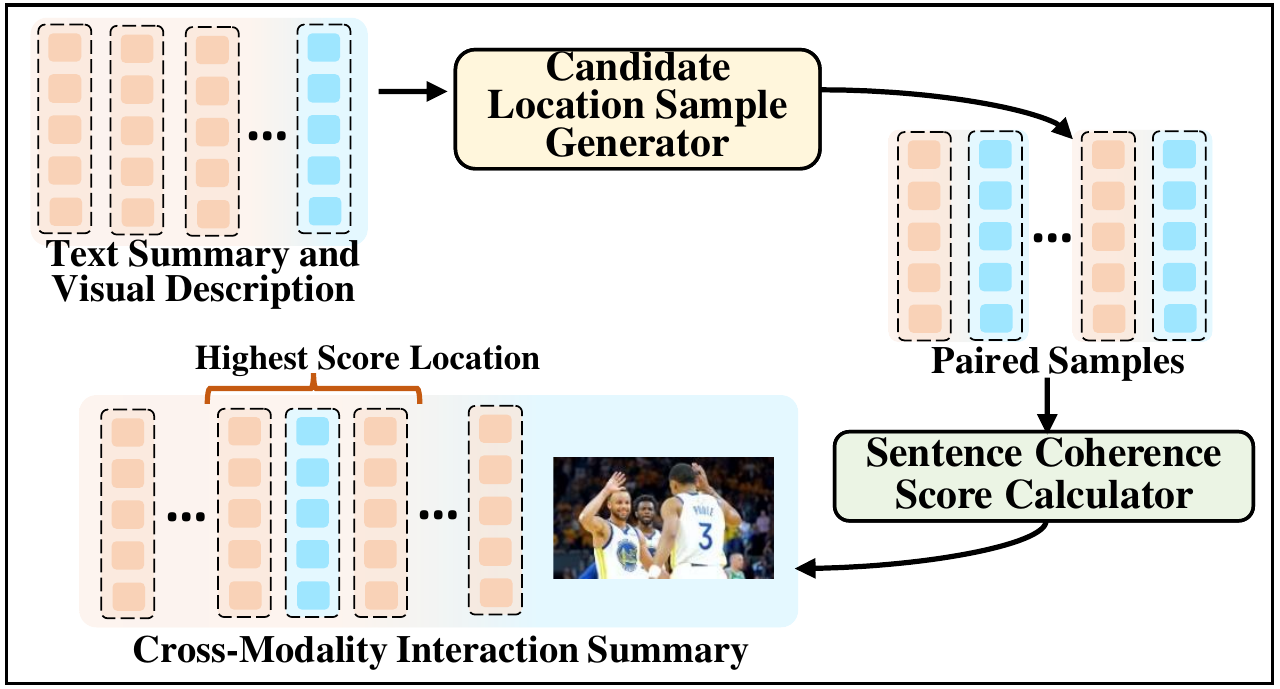}
    \end{center}
    \caption{The overview of Visual Description Locator.}
    \label{fig:VDL}
\end{figure}

After the Multi-task Triple Decoder, a Visual Description Locator (VDL) is designed to combine the text summary and the relevant visual description.
As shown in Figure \ref{fig:VDL}, the VDL has two core modules. Candidate Location Sample Generator splits the text summary into short sentences through punctuation and combines them with visual description. And then, the paired samples are fed into Sentence Coherence Score Calculator (SCSC) to calculate location scores. The highest score is the most coherence location to insert the visual description. 

Especially, SCSC is trained as a BERT-based \cite{2018BERT} sequence classification model with self-supervised data from the source article. We randomly select two continuous short sentences from the article title for positive samples and select two discontinuous sentences from the same title or cross-article for the negative samples. Finally, we generated 980k samples to finetune SCSC, and the test accuracy arrived at 0.98. 
\section{Experiment Setup}

\subsection{Datasets}
We evaluate our model on a Chinese multimodal summarization dataset \cite{Zhang2021HierarchicalCS}. 
The text data are composed of various types of news from TTNews \cite{hua2017overview}, and THUCNews \cite{sun2016thuctc}, including sports, technology, entertainment, society, etc. 
Each article contains the relevant images with captions, and the most relevant image is annotated.
The training, validation, and test sets contain 52656, 5154, and 5070 samples. The articles and summaries have 955.26 and 36.61 tokens on average, respectively.

\subsection{Baselines}
Three categories of baselines are compared.\par

\noindent\textit{{Multimodal Summarization with Multimodal Output}}: \noindent\textbf{MSMO} \cite{zhu2018msmo} firstly proposes multimodal output summarization, in which the attention mechanism is used to generate textual summaries.
\noindent\textbf{MOF} \cite{zhu2020multimodal} is an MSMO-based model that includes image accuracy as an additional loss.
\noindent\textbf{VMSMO} \cite{li2020vmsmo}
proposes a Dual-Interaction Multimodal Summarizer to generate multimodal output summarization. 
\noindent\textbf{HCSCL} \cite{Zhang2021HierarchicalCS} is a hierarchical cross-modality semantic correlation learning model to learn the intra- and inter-modal correlation. \par

\noindent\textit{{Multimodal Summarization with Text Output}}: \noindent\textbf{MAtt} \cite{li2018multi} is an attention-based model that constructs an image content filter to combine visual features and text. \noindent\textbf{HOW2} \cite{palaskar2019multimodal} is the first model proposed to summarize the video content with textual summary. \noindent\textbf{MSE} \cite{li2020multimodal} is a model that utilizes a selective gate network to improve the encoder's capacity to recognise news highlights. \par

\noindent\textit{{Traditional Text-only Summarization}}:
\noindent\textbf{PG} \cite{see2017get} is a sequence-to-sequence framework with attention mechanism and pointer network. \noindent\textbf{S2S} \cite{luong2015effective} is a standard sequence-to-sequence architecture using a RNN encoder-decoder with a global attention mechanism. 
\subsection{Textual Overlap Metrics}
As the standard metric in the field of text generation, ROUGE \cite{lin2004rouge}and BLEU \cite{papineni2002bleu}  are applied to evaluate the overlap between the generated summary and the target text. For relevant image selection ,an image precision (IP) score \cite{zhu2018msmo} is used to evaluate whether an image is correctly selected as output.

\begin{table*}[!ht]
\setlength{\abovecaptionskip}{0.2cm} 
\setlength{\belowcaptionskip}{-0.2cm} 
  \centering
  \renewcommand\arraystretch{0.85}
    \begin{tabular}{lrrrrrrrr}
    \toprule
          & \multicolumn{1}{c}{R-1} & \multicolumn{1}{c}{R-2} & \multicolumn{1}{c}{R-L} & \multicolumn{1}{c}{B-1} & \multicolumn{1}{c}{B-2} & \multicolumn{1}{c}{B-3} & \multicolumn{1}{c}{B-4} & \multicolumn{1}{c}{IP} \\
    \midrule
    \textit{Traditional Textual Model} &       &       &       &       &       &       &       &  \\
    PG\cite{see2017get}    & 42.54 & 28.25 & 39.95 & 38.26 & 30.78 & 24.55 & 19.92 & \multicolumn{1}{c}{-} \\
    S2S\cite{luong2015effective}   & 30.13 & 14.40  & 28.61 & 26.36 & 17.55 & 11.07 & 7.44  & \multicolumn{1}{c}{-} \\
    \midrule
    \multicolumn{9}{l}{\textit{Multimodal Summarization} Model} \\
    Matt\cite{li2018multi}  & 41.30  & 24.70  & 37.55 &  \textbf{37.41} & 28.44 & 21.13 & 16.02 & \multicolumn{1}{c}{-} \\
    HOW2\cite{palaskar2019multimodal}  & 39.53 & 20.28 & 35.55 & 35.44 & 24.42 & 17.81 & 13.71 & \multicolumn{1}{c}{-} \\
    MSE\cite{li2020multimodal}   & 42.99 & 28.78 &  \textbf{41.79} & 39.02 & 31.51 & 25.21 & 20.50  & \multicolumn{1}{c}{-} \\
    \midrule
    \multicolumn{9}{l}{\textit{Multimodal Summarization Output Model}} \\
    MSMO\cite{zhu2018msmo}  & 42.89 & 28.25 & 39.86 & 39.06 & 31.26 & 24.78 & 19.93 & 33.50 \\
    MOF\cite{zhu2020multimodal}   & 42.41 & 28.10  & 39.60  & 38.04 & 30.55 & 24.38 & 19.71 & 33.16 \\
    VMSMO\cite{li2020vmsmo} & 42.68 & 28.35 & 41.34 & 38.75 & 31.20  & 24.75 & 19.99 & 32.86 \\
    HCSCL\cite{Zhang2021HierarchicalCS} &  \textbf{43.64} & 29.00    & 40.94 & 39.64 & 31.91 & 25.40  & 20.54 & 40.98 \\
    \midrule
    \multicolumn{9}{l}{\textit{Our Model}} \\
    CISum &43.21       & \textbf{30.39}       &40.51       &39.02       &  \textbf{32.20}      & \textbf{26.54}       & \textbf{22.55}       & \textbf{73.76}  \\
    \bottomrule
    \end{tabular}%
  
    \caption{Rouge, Bleu and IP scores comparison with
  summarization baselines.}
  \label{table:table1}
\end{table*}%
\subsection{Multimodal Semantics Coverage Metric}
We also design the multimodal semantics coverage metric to evaluate the multimodal similarity between the MS output and the target summary. To measure cross-modality similarity, the visual and textual representation should be mapped into the same semantic space. The multimodal pretrained model CLIP \cite{Radford2021LearningTV} is applied to extract different modalities' features to calculate the multimodal similarity. The CISum's multimodal semantics is formulated as follows:
\begin{equation}
\begin{split}
    Y_t^{\prime},Y_d^{\prime} &= F_{split}(Y_t,Y_d) \\
    H_{E}&=F_{CLIP}(Y_t^{\prime},Y_d^{\prime},Y_v) \\
\end{split}
\end{equation}
where $Y_t,Y_v$ and $Y_d$ denote the text summary output, image selection result and visual description output. $F_{split}$ is the text content split function to split long sentences into short ones by punctuation. $F_{CLIP}$ is the CLIP function to map different modalities into the same space. $H_{E} = \{h_1^{E},h_2^{E},...,h_n^{E} \}$ denotes the multimoal feature of CISum, and $H_{T} = \{h_1^{T},h_2^{T},...,h_n^{T}\}$ denotes the multimodal feature of the target summary. Then multimodal semantics coverage metric MSC-P, MSC-R and MSC-F1 are defined as follows:
\begin{equation}
\begin{split}
\text{MSC-P} = \frac{\mid F_{match}(H_{E},H_{T})\mid}{\mid H_{E} \mid}\\
\text{MSC-R} = \frac{\mid F_{match}(H_{E},H_{T})\mid}{\mid H_{T} \mid}\\
\text{MSC-F1} = 2*\frac{\text{MSC-P}*\text{MSC-R}}{\mid \text{MSC-P}+\text{MSC-R} \mid}\\
\end{split}
\end{equation}
where $F_{match}$ is to select the coveraged semantic feature in the target summary $H_T$ based on multimodal summarization result $H_E$. For every feature $h_i^E$, $F_{match}$ calculates the cosine similarity score and the feature with the highest score in $H_E$ is selected, which is denoted as $\arg\max CosSim(h_1^{E},H_{T})$. 

\subsection{Implementation Details}
For the model shown in Figure 2, we set $L=P=2$, $K=D=M=6$, and the multi-head number of all transformer layers is set to 8. In the text embedding module, we load the word embedding of BERT \cite{2018BERT} to initialize our embedding matrix, which has a size of 21128 and a dimension of 768. The image feature extracted by a bottom-up attention model \cite{2017Bottom} with confidence larger than 0.55 is selected. Each object is represented by a 2048-dimensional vector. We employ Adam optimizer \cite{2014Adam} with the initial learning rate $2e\text{-}4$ multiplied by 0.9 every 10 epochs. 
The experiments are conducted on an NVIDIA Tesla V100 GPU.
\section{Results and Analysis}

\subsection{Text Semantic Coverage Analysis}

Tabel \ref{table:table1} shows the traditional ROUGE and BLEU scores of nine baseline models and CISum on the test dataset. Although text summary generation is only one subtask, CISum is still comparable to the most advanced MS model. Specifically, CISum achieves the highest scores of ROUGE-1, BLEU-2, BLEU-3, and BLEU-4. It indicates that the multi-task learning framework can generate the text summary with high semantics coverage, and the other tasks also can improve the text summarization performance.

Furthermore, Table \ref{table:table1} also shows the result of IP, which evaluates the performance of relevant image selection. CISum achieves the highest IP score, which is higher than the best baseline by 32.78. It demonstrates the superiority that the cross-modality multi-task learning framework is more effective in learning the semantics of visual content.

\begin{table}[!t]
\setlength{\abovecaptionskip}{0.2cm} 
\setlength{\belowcaptionskip}{-0.4cm} 
  \centering
  \renewcommand\arraystretch{0.85}
  
    \begin{tabular}{lrrr}
    \toprule
          & \multicolumn{1}{c}{MSC-R} & \multicolumn{1}{c}{MSC-P} & \multicolumn{1}{c}{MSC-F1} \\
    \midrule
    PG    & 39.55  & 75.29  & 51.86  \\
    S2S   & 35.60  & 67.25  & 46.56  \\
    Matt  & 37.62  & 78.59  & 50.88  \\
    HOW2  & 38.49  & 74.68  & 50.80  \\
    MSMO  & 37.04  & 81.67  & 50.97  \\
    VMSMO & 35.97  & 82.07  & 50.02  \\
    \midrule
    CISum & \textbf{54.82}  & 66.82  & \textbf{60.23}  \\
    \bottomrule
    \end{tabular}%
    \caption{Multimodal semantics coverage results.}
    \label{table:table2}
\end{table}%

\begin{table*}[!ht]
\setlength{\abovecaptionskip}{0.2cm} 
\setlength{\belowcaptionskip}{-0.2cm} 
  \centering
  \renewcommand\arraystretch{0.7}
   \resizebox{\textwidth}{!}{
    \begin{tabular}{lccccccccccc}
    \toprule
          & R-1   & R-2   & R-L   & IP    & I-R-1 & I-R-2 & I-R-L & I-B-1 & I-B-2 & I-B-3 & I-B-4 \\
    \midrule
    w/o Filter & 42.63  & 29.47  & 40.36  & 62.49  & 25.42  & 16.54  & 25.33  & 19.09  & 14.08  & 10.72  & 8.91  \\
    w/o Img-S & 41.34  & 28.21  & 38.40  & -     & 23.38  & 14.31  & \textbf{27.59 } & 21.00  & 15.29  & 11.49  & 9.54  \\
    w/o Desc-D & 42.99  & 30.27  & 40.24  & 50.94  & -     & -     & -     & -     & -     & -     & - \\
    CISum & \textbf{43.21 } & \textbf{30.39 } & \textbf{40.51}  & \textbf{73.76 } & \textbf{25.97 } & \textbf{17.29 } & 26.51  & \textbf{23.37 } & \textbf{17.98 } & \textbf{14.20 } & \textbf{12.09 } \\
    \bottomrule
    \end{tabular}%

  }
  \caption{Ablation results. "w/o Filter" means without the noisy filter module. "w/o Img-S" means without relevant image selector. "w/o Desc-D" means without visual description decoder. I-R-1, I-R-2, I-R-L mean visual description ROUGE scores, and I-B-1, I-B-2, I-B-3, I-B-4 mean visual description BLEU scores.}
 \label{table:table3}
\end{table*}%
\begin{figure*}[!th]
\setlength{\abovecaptionskip}{0.05cm} 
\setlength{\belowcaptionskip}{-0.2cm} 
    \begin{center}
    \includegraphics[scale=0.85]{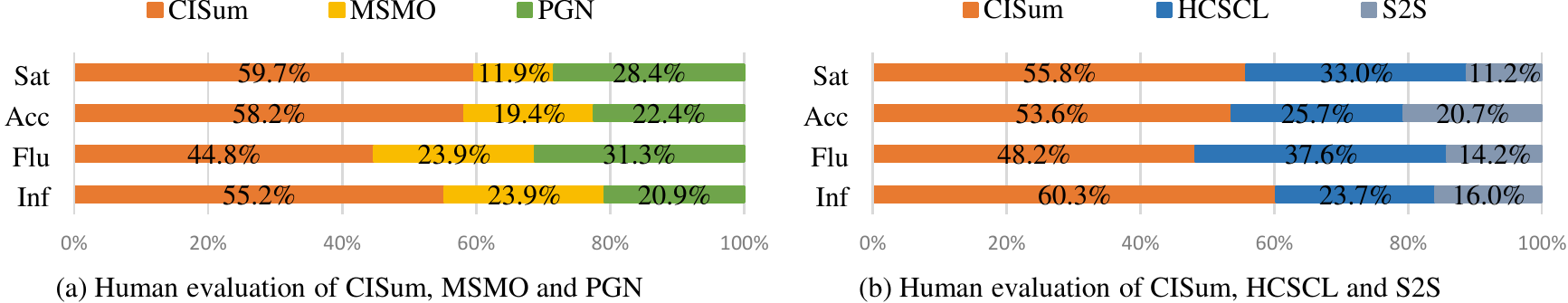}
    \end{center}
    \caption{Human evaluation results of baselines.}
    \label{fig:human}
\end{figure*}

\begin{figure*}[!h]
\setlength{\abovecaptionskip}{0.05cm} 
\setlength{\belowcaptionskip}{-0.2cm} 
    \begin{center}
    \includegraphics[scale=0.85]{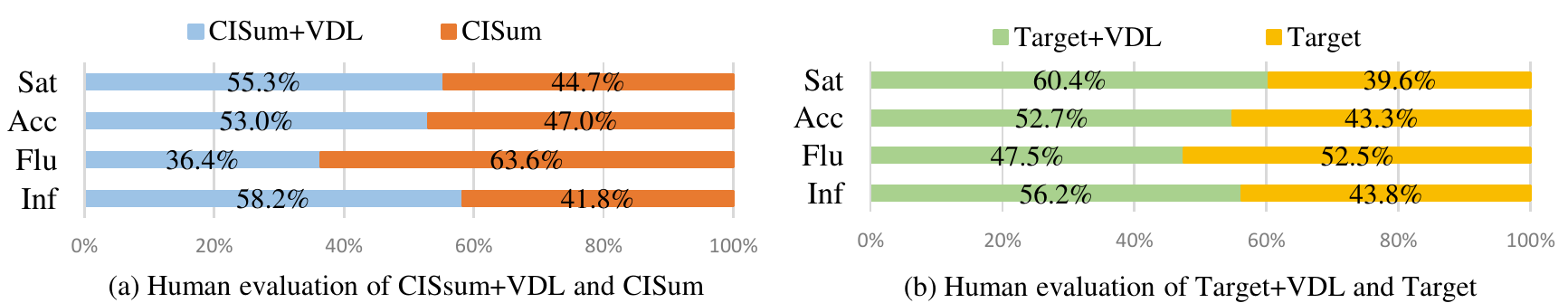}
    \end{center}
    \caption{Human evaluation results of VDL.}
    \label{fig:vdl_result}
\end{figure*}
\subsection{Multimodal Semantics Coverage Analysis}
As discussed above, the output of different modalities in the MS models should be mapped into the same feature space to evaluate the multimodal semantics coverage. As mentioned in 4.3, we apply the pretrained model CLIP to embed the multimodal features and compare the output with the target to calculate the multimodal semantics coverage metric. Table \ref{table:table2} shows the result of MSC-R, MSC-P, and MSC-F1. We can see that CISum outperforms baselines 15.27 and 8.37 in terms of MSC-R and SMC-F1. It demonstrates that combining text summary, selected image, and image description by CISum can cover the target multimodal semantics more comprehensively. As for MSC-P, CISum is slightly worse. After analyzing the cases, we find that the semantic content of CISum is more abundant than baselines, and the human evaluation in section 5.4 also shows that CISum achieves higher informativeness scores. The extra semantics information decreases the MSC-P score, but MSC-F1 is still high due to the high semantics coverage.
\subsection{Ablations Study}
Three ablation experiments are designed to verify the effectiveness of the noisy filter module, relevant image selector, and visual description decoder. The result is shown in Table \ref{table:table3}. 
The visual description ROUGE and BLEU scores are higher with the noisy filter module. 
It demonstrates that the noisy filter module reduces textual noise effectively and improves visual representation.
With the visual description decoder, the IP score is higher. 
With the relevant image selector, the visual description ROUGE and BLEU scores are higher. 
It indicates that cross-modality multi-task learning improves the performance of all the sub-tasks.

\subsection{Human Evaluation}
We further conduct the human evaluation to analyze the summary output from four aspect metrics, i.e., informativeness (Inf)—whether the summary provides enough and necessary information of the input, fluency (Flu)—whether the summary is grammatically correct, accuracy (Acc)—whether the summary content is accurate or confused, satisfaction (Sat)—whether the reading experience is satisfied. 100 examples of the summarization result are randomly selected, and three annotators volunteer to choose the best one from the comparison methods based on the four metrics.

The comparison of PG, MSMO, and CISum is shown in Figure \ref{fig:human} (a), in which CISum achieves the highest percentage scores in all metrics. Especially in Sat, Acc, and Inf, CISum arrives 59.7\%, 58.2\%, and 55.2\%. The comparison of S2S, HCSCL, and CISum is shown in Figure \ref{fig:human} (b). CISum also achieves the highest percentage scores, especially in the metrics of Sat and Inf.  
It demonstrates that CISum can enrich the summary output and improve visual information accuracy.

\subsection{Visual Description Locator Analysis}

The same as human evaluation, four aspect metrics, i.e., Inf, Flu, Acc, and Sa), are applied to evaluate the effectiveness of Visual Description Location (VDL). 100 examples are randomly selected, and three annotators are asked to choose the best one.
As shown in Figure \ref{fig:vdl_result} (a), with the VDL module, Sat and Inf scores are higher. It indicates that inserting the visual description into the textual summary helps the reader acquire more straightforward and informative information to a certain degree. With the VDL module, the Flu score is slightly lower. The reason is that the inserted sentence may decrease context fluency. To further explore the human preference for output summarization, we also evaluate the target and target+VDL results. Figure \ref{fig:vdl_result} (b) shows that the Sat and Inf are higher with the VDL. The above experiments show that the VDL module improves multimodal summary output informativeness and readability, which is effective for the entire reading experience.
\subsection{Case Study}
\begin{figure}[!h]
\setlength{\abovecaptionskip}{0.05cm} 
\setlength{\belowcaptionskip}{-0.6cm} 
    \begin{center}
    \includegraphics[scale=0.92]{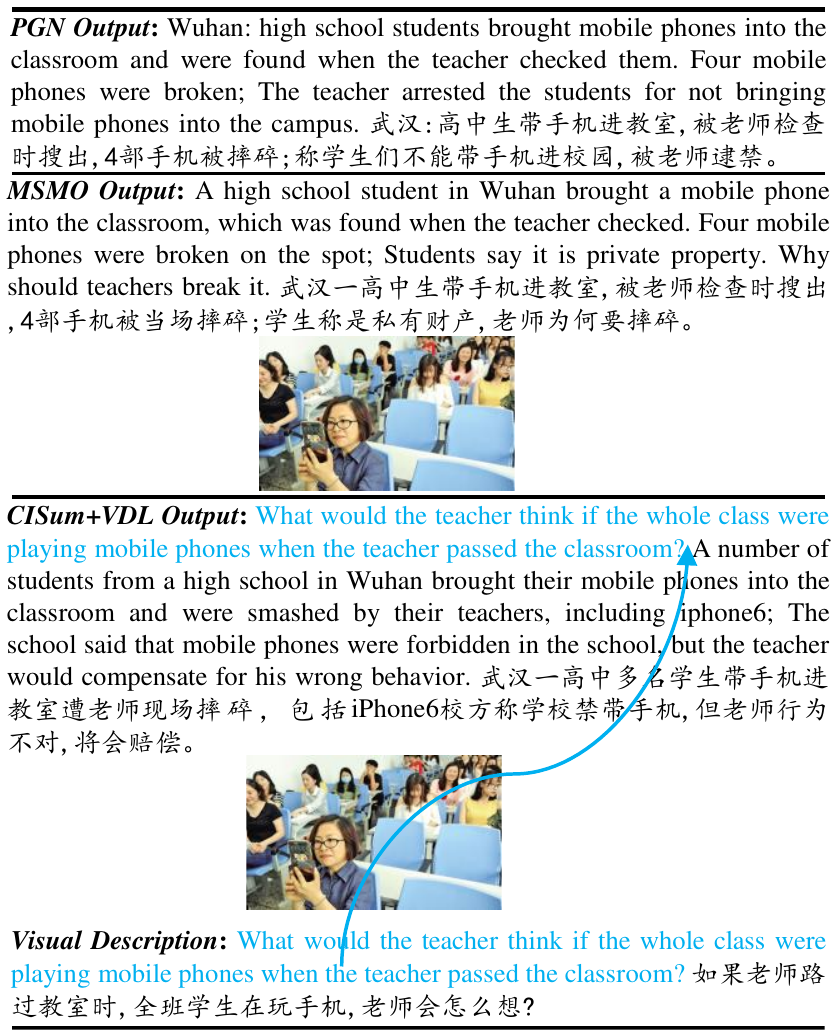}
    \end{center}
    \caption{Summary output comparison between CISum+VDL and baselines.}
    \label{fig:case_study1}
\end{figure}

Case studies are shown in Figure \ref{fig:case_study1}, which shows the cross-modality interaction summaries generated by CISum. We also show the original article and reference summary. Compared with text-only PGN output, MSMO output with a selected image contains more information. But a contextually irrelevant image may cause ambition and confuse readers. The CISum generates a background-based visual description and inserts it into the appropriate location by VDL. The interaction summary output achieves higher multimodal semantic coverage.
\section{Related work}
According to the format of inputs, summarization tasks can be divided into text-only summarization and multimodal summarization.

Text-only summarization aims to generate the most salient information from the textual input. ABS \cite{rush2015neural} is the first model to generate summarization. S2S \cite{luong2015effective} proposes a global-local approach to improve effectiveness. PG \cite{see2017get} consists of the Seq2Seq model, pointer generation network, and an overwrite copy mechanism. Some pretrained methods \cite{DBLP:conf/icml/ZhangZSL20, qi-etal-2020-prophetnet} have recently have been proposed, which achieved better performance of overlap metric. These methods are only propose to handle the text input.

Multimodal summarization aims to condense information from multimodal inputs, such as text, vision, and audio \cite{li2017multi}. 
Recently, Ms has been extensively studied (\cite{Li2020AspectAwareMS,chen2018abstractive,khullar2020mast,zhu2020multimodal}). A large number of the works focus on fusing visual information to improve the quality of text summaries.
\cite{li2018multi} introduces a multimodal sentence summarization task that creates a condensed text summary from a long sentence and the corresponding image. 
\cite{palaskar2019multimodal} propose a multimodal attention model for How2 videos \cite{sanabria2018how2}. 
\cite{li2020multimodal} design a gate to select event highlights from images and distinguishes highlights in encoding. 
\cite{li2020vmsmo} propose a local-global attention mechanism to let the video and text interact and select an image as output. 
These works mainly consider the visual content as the supplement to improve textual semantics coverage. We are the first work to focus on improving multimodal semantics coverage improvement by exploiting the interaction between different modalities.

\section{Conclusion}
This paper proposes a multi-task cross-modality learning framework to learn the cross-modality interaction between text and image, which effectively improves the multimodal semantics coverage of summary. Experimental results show that the generated summary is more fluent, informative, relevant, and less confused while maintaining the excellent performance of word overlap metrics. The novelty of this work is to fuse the two modalities as a whole to generate a summary capturing the semantics of different modalities. It is different from existing works that independently handle different modalities and consider visual content as a supplement to improve text summary. We complement current researches that are mainly extended from the text-only summarization methods, which are not effective to learn the multimodal semantics.
\section{Acknowledgements}
This research is supported by the National Natural Science
Foundation of China (No.62272025 and No.U22B2021), and
the Fund of the State Key Laboratory of Software Development Environment.
\bibliography{custom_2}
\bibliographystyle{siam}
\end{document}